\documentclass{article} %

\usepackage{colm2024_conference}

\usepackage{booktabs}
\usepackage{graphicx}
\usepackage{enumitem}
\usepackage{wrapfig}
\usepackage{algorithm}
\usepackage{algpseudocode}
\usepackage[T1]{fontenc}

\usepackage{lmodern}
\usepackage{pifont}

\usepackage{microtype}
\usepackage{amsmath}
\usepackage{colortbl}
\usepackage[utf8]{inputenc}
\definecolor{lightgray}{rgb}{0.9,0.9,0.9}
\usepackage{caption}
\usepackage{subcaption}
\usepackage{xcolor}
\usepackage{setspace}
\usepackage{url}
\usepackage{multirow}
\usepackage{colortbl}
\usepackage{tabularx}
\usepackage{blindtext}
\usepackage{pgfplots}
\pgfplotsset{compat=1.18} 
\usepackage{tikz}
\usetikzlibrary{er,positioning,bayesnet}
\usepackage{makecell}
\usepackage{tipa}
\usepackage{siunitx}
\usepackage{nicefrac}
\usepackage{tocloft}
\usepackage{listings}
\usepackage[raster,skins]{tcolorbox} %
\usepackage{xltabular}
\usepackage{adjustbox}
\usepackage{xurl}
\usepackage{marvosym}
\usepackage{algpseudocode}

\usepackage{amsmath}
\usepackage{amssymb}
\usepackage{mathtools}
\usepackage{amsthm}
\usepackage{enumitem}

\usepackage[table]{xcolor}

\definecolor{outlier1}{RGB}{250,252,255}
\definecolor{outlier2}{RGB}{238,244,250}
\definecolor{outlier3}{RGB}{226,236,245}
\definecolor{outlier4}{RGB}{210,225,240}
\definecolor{outlier5}{RGB}{190,210,230}
\definecolor{outlier6}{RGB}{170,195,220}
\definecolor{outlier7}{RGB}{150,180,210}

\definecolor{loss1}{RGB}{250,253,250}
\definecolor{loss2}{RGB}{242,250,240}
\definecolor{loss3}{RGB}{230,245,230}
\definecolor{loss4}{RGB}{215,235,210}
\definecolor{loss5}{RGB}{200,225,200}
\definecolor{loss6}{RGB}{185,215,185}
\definecolor{loss7}{RGB}{170,205,170}

\definecolor{gap_red1}{RGB}{255,245,205}
\definecolor{gap_red2}{RGB}{255,230,190}
\definecolor{gap_red3}{RGB}{255,210,170}
\definecolor{gap_green1}{RGB}{235,250,205}
\definecolor{gap_green2}{RGB}{220,245,190}
\definecolor{gap_green3}{RGB}{200,235,170}

\author{\textbf{Zihan Qiu}$^*$$^{1}$, \textbf{Zeyu Huang}$^*$$^{2}$, \textbf{Kaiyue Wen}$^*$$^{3}$, \textbf{Peng Jin}$^*$$^{1}$,  \textbf{Bo Zheng}$^*$$^{1}$,  \\   \textbf{Yuxin Zhou}$^{1}$, \textbf{Haofeng Huang}$^{1,4}$, \textbf{Zekun Wang}$^{1}$, \textbf{Xiao Li}$^{1}$, \textbf{Huaqing Zhang}$^{1,4}$, \\ \textbf{Yang Xu}$^{1}$, \textbf{Haoran Lian}$^{1}$, \textbf{Siqi Zhang}$^{1}$, \textbf{Rui Men}$^{1}$, \textbf{Jianwei Zhang}$^{1}$, \\ \textbf{Ivan Titov}$^{2}$, \textbf{Dayiheng Liu}$^\dagger$$^{1}$, \textbf{Jingren Zhou}$^{1}$, \textbf{Junyang Lin}$^\dagger$$^{1}$
\\$^1$Qwen Team $^2$University of Edinburgh $^3$Stanford University  $^4$Tsinghua University \\
\small $^*$Equal contribution.\quad $^\dagger$Corresponding authors.
}
\title{A Unified View of Attention and Residual Sinks: \\Outlier-Driven Rescaling is Essential for Transformer Training}

\begin{document}

\maketitle

\begin{abstract}
We investigate the functional role of emergent outliers in large language models, specifically attention sinks (a few tokens that consistently receive large attention logits) and residual sinks (a few fixed dimensions with persistently large activations across most tokens).
We hypothesize that these outliers, in conjunction with the corresponding normalizations (\textit{e.g.}, softmax attention and RMSNorm), effectively rescale other non-outlier components.
We term this phenomenon \textit{outlier-driven rescaling} and validate this hypothesis across different model architectures and training token counts.
This view unifies the origin and mitigation of both sink types.
Our main conclusions and observations include:
(1) Outliers function jointly with normalization: removing normalization 
eliminates the corresponding outliers but degrades training stability and performance; directly clipping outliers while retaining normalization leads to degradation, indicating that outlier-driven rescaling contributes to training stability.
(2) Outliers serve more as rescale factors rather than contributors, as the final contributions of attention and residual sinks are significantly smaller than those of non-outliers.
(3) Outliers can be absorbed into learnable parameters or mitigated via explicit gated rescaling, leading to improved training performance (average gain of 2 points) and enhanced quantization robustness (1.2 points degradation under W4A4 quantization).
\end{abstract}

\section{Introduction}

Transformer-based Large Language Models (LLMs) exhibit outliers. 
These extreme values, exceeding regular activations or logits by orders of magnitude, pose practical challenges: they dominate the dynamic range of the representations during model quantization~\citep{yao2022zeroquant,xiao2023efficient,xiao2023smoothquant,wei2023outlier,nvfp4}, and could lead to larger numerical errors in floating-point arithmetic~\citep{budzinskiy2025numerical}. 
However, simply removing them through clipping severely degrades model performance~\citep{kovaleva2021bert,sun2024massive}, suggesting that they play an essential functional role in transformers. 

A prominent instance of outliers is the \textit{attention sink}~\citep{xiao2023efficient}, where a small subset of attention logits becomes larger than the rest, causing a few special tokens (sink tokens) to consistently receive high attention scores.
Recent work reveals that their formation is intrinsically linked to softmax normalization~\citep{bondarenko2023quantizable,an2025systematic}, and their corresponding value vectors exhibit significantly smaller norm than those of non-sink tokens~\citep{sun2024massive,an2025systematic}, indicating that these sink tokens do not dominate the attention output with their abnormally large attention score, but leverage it as the scaling factor within softmax normalization in attention.
This view is further supported by the introduction of GatedAttention (GA) ~\citep{bondarenko2023quantizable,qiu2025gated,an2025systematic}, where an explicit gating mechanism enables the model to perform such rescaling, thereby mitigating the reliance on attention sinks.
Another notable outlier phenomenon is massive activation (MA)~\citep{sun2024massive} in the residual stream: tokens associated with attention sinks often exhibit extremely large activations in specific dimensions, which, after passing through normalization layers, can promote the formation of attention sinks \citep{an2025systematic}. 

\begin{figure}[ht]
    \centering
    \vskip  -0.3in
    \includegraphics[width=\linewidth]{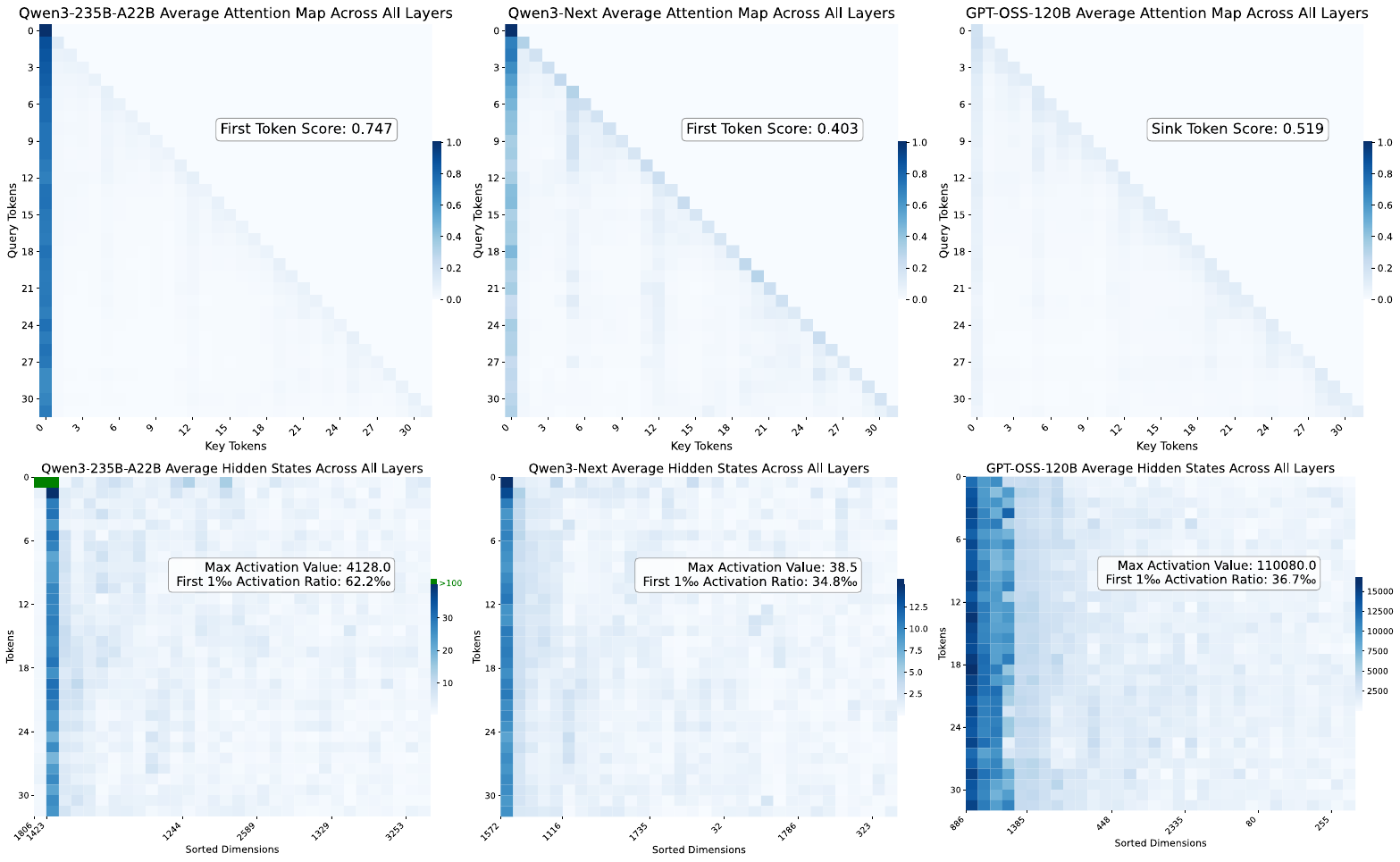}
    \vskip  -0.05in
    \caption{\footnotesize In the first row, all models exhibit varying degrees of attention sinks: the first token produces attention logits significantly larger than those of other tokens, dominating the attention scores.
    In the second row, Qwen3-235B-A22B shows massive activations, with dimensions 1806 and 1423 of the first token exceeding 1000. 
    Beyond these extreme values, all models display consistent residual sinks: certain fixed dimensions yield persistently higher activations across all tokens compared to others.
    }
    \vskip  -0.15in
    \label{fig:all_sinks}
\end{figure}

Both types of outliers above share a key characteristic: they exert their effects via normalization. We unify this behavior under the term \textit{outlier-driven rescaling}, in which outliers interact with normalization to rescale the non-outlier components after normalization. 
We validate the outlier-driven rescaling hypothesis on a distinct type of outliers within the residual~\citep{dettmers2022gpt3,bondarenko2023quantizable}.
These outliers appear in a fixed set of dimensions across the vast majority of tokens, exhibiting activations that are orders of magnitude larger than typical values, as shown in Fig.~\ref{fig:all_sinks}.
Our experiments show that these outliers share many properties with attention sinks, which motivates us to name them \textit{residual sinks}. 
Notably, residual sinks are not tied to specific inputs, suggesting they are not passing data-specific features.
We further verify that they interact with the RMSNorm layers~\citep{zhang2019root} to perform outlier-driven rescaling. 
Extensive empirical evidence supports this hypothesis, as detailed below.

\textbf{(1)} Removing normalization~\citep{zhu2025transformers} reduces residual sinks but hurts model performance and training stability~(Sec.\ref{sec:dyt}). 
Suppressing outliers via clipping or architectural modifications also yields degradation (Sec.~\ref{sec:clip}).

\textbf{(2)} We observe that in models that rely on outlier-driven rescaling, the RMSNorm weights for outlier-prone dimensions are consistently much smaller than the mean (e.g., 0.006 vs. 1), further suggesting that these outlier dimensions act primarily as rescale factors rather than direct contributors to the normalized output.
\textit{We prove the upper bound on the feature norm after RMSNorm decreases as the outlier magnitude increases given this property }(App.~\ref{proof}).

\textbf{(3)} Residual sink can be absorbed in parameters, analogous to the absorption of attention sinks into learnable biases~\citep{sun2024massive,agarwal2025gpt}, eliminating the need for explicit outliers in the residual stream~(Sec.\ref{sec:pre_affine}).

\textbf{(4)} If rescaling is the underlying purpose, explicitly introducing it is expected to reduce outliers.
Consistent with prior work showing that gating-based rescaling in attention reduces attention sinks~\citep{bondarenko2023quantizable,an2025systematic,qiu2025gated}, our experiments demonstrate that inserting a lightweight gating after RMSNorm (GatedNorm) effectively mitigates residual sinks while preserving or even enhancing model performance (Sec.~\ref{sec:gate}).
With fewer outliers and smoother activations, the model trained with GatedNorm exhibits better quantization performance.

\textbf{(5)} Gating-based rescaling reduces the model’s reliance on outliers, thereby weakening sensitivity to outlier-inducing architectural choices. 
For example, while SwiGLU~\citep{shazeer2020glu} typically outperforms sigmoid-based GLU in FFNs by generating larger activations that support outlier-driven rescaling, adding gating-based rescaling enables GLU to match or exceed SwiGLU (Fig.~\ref{fig:swiglu}).

We validate our hypothesis across a wide range of models, including standard softmax attention~\citep{vaswani2017attention}, linear attention~\citep{yang2024gated}, linear-full attention hybrid architectures with 1B, 7B, and 24B parameters, trained on datasets ranging from 120B to 1T tokens. 
In the App.~\ref{app:comparision}, we summarize a comparison of attention sinks and residual sinks under the outlier-driven rescaling perspective, covering their patterns, functional roles, and mitigations.
Our results consistently indicate that outliers at normalizations in softmax attention and RMSNorm are not pathological artifacts but rather rescaling-driven factors.
By understanding the functional role of the residual sink, we propose simple, effective alternatives that reduce residual sink while preserving their functional benefits, thereby improving training and quantization performance.

\section{Outliers in Large Language Models}

In this section, we present outliers in open-source LLMs and examine their interrelations. 
We focus on pre-norm transformers.
For an input sequence of length \(L\), the model first embeds tokens into hidden states \(\mathbf{H_0} \in \mathbb{R}^{L \times d}\), where \(d\) is the hidden dimension. The states are then processed through \(D\) layers. Denoting the \(i\)-th layer function as \(F_i\), the hidden states are updated as:
\[
\mathbf{H}_{i+1} = \mathbf{H}_i + F_i(\mathbf{H}_i), \quad \text{for } i = 0, 1, \dots, D-1.
\]
The hidden states \(\mathbf{H}_i\), also referred to as the residual stream, constitute the primary focus of our study.

We analyze four models with distinct outlier patterns: Qwen3-235B-A22B~\citep{yang2025qwen3}, Qwen3-Next~\citep{yang2025qwen3}, GPT-OSS~\citep{agarwal2025gpt}, and DeepSeek-V3~\citep{liu2024deepseek} in Fig.~\ref{fig:all_sinks}. 
For each model, we use the same set of input sequences and record both the attention maps and the residual activations across all layers.
In Fig.~\ref{fig:all_sinks}, we average attention maps and hidden states over all layers for clarity.
Because the hidden dimension \(d\) is large, visualizing all dimensions is impractical. 
To highlight outlier patterns, we reorder the feature dimensions by their overall activation magnitude. 
Specifically, for each dimension \(j\), we compute its average absolute activation across all tokens, layers, and inputs:
\[
\mathbf{H}^j_{\text{avg}} = \frac{1}{N(D+1)} \sum_{n=0}^{N-1} \sum_{i=0}^{D} |\mathbf{H}^j_{i,n}|,
\]
where \(\mathbf{H}^j_{i,n}\) is the activation of the \(j\)-th dimension for the \(n\)-th token at the \(i\)-th layer, and \(N\) is the number of tokens. 
We then sort dimensions in descending order of \(\mathbf{H}_{\text{avg}}\).
After this reordering, dimensions with consistently large activations appear on the left side of our visualizations, forming a structure that closely resembles the attention sink pattern.

In Qwen3-235B-A22B, the first token consistently receives high attention scores from nearly all other tokens, exhibiting \textit{attention sink}.
Correspondingly, this token shows MA in two dimensions (e.g., dimension 1806 and 1423). 
Beyond token-specific MA, we observe that \textit{most} tokens exhibit consistently large activations in dimensions 1423.
The pattern is stable across inputs, appearing as a dark vertical stripe in activation visualization, similar to shape of the attention sinks. 
We later find that its rescaling effect during normalization closely resembles that of the attention sink, leading us to term this \textit{residual sink}.
A similar pattern is observed in Deepseek-V3 in App.~\ref{fig:reduce_sink_dpsk}: attention sinks, MA, and residual sinks all exist.

Qwen3-Next introduces a gating in the attention to perform gating-based rescaling, thereby reducing its reliance on attention logit outliers.
As a result, attention sinks are weaker than others. 
Moreover, the maximum activation magnitude in its residual stream is only 38.5 and no prominent MA are observed. 
Nevertheless, dimension 1572 consistently exhibits large activations higher than all other dimensions across tokens, clearly manifesting a residual sink.

GPT-OSS incorporates learnable sinks, effectively removing attention sinks from real input tokens. 
MA also disappear: the hidden states of real tokens no longer exhibit extreme values in specific dimensions.
This aligns with prior interpretation: attention sinks act as an input-independent bias in softmax attention; MA enable this by producing near one-hot vectors after normalization, which activate only a few fixed matrix columns when projected into the key space.
When a learnable bias (e.g., a dedicated sink key) is provided explicitly, the model no longer needs to generate MA from real tokens.
However, despite the absence of attention sinks and MA, residual sinks persist in GPT-OSS. 

\section{Outlier-Driven Rescaling}

In this section, we provide a detailed discussion and empirical evidence on the roles of these outliers. 
We conduct our experiments on the pre-norm transformer, closely following the design of dense models in Llama3~\citep{dubey2024llama} and Qwen3~\cite{yang2025qwen3}. 
Due to the large number of ablation and comparison conditions, we primarily evaluate all variants under a consistent setting: a 2B-parameter model trained on 120B tokens.
When structural changes affect the model parameters, \textit{we adjust the FFN width accordingly to maintain a constant total parameter count}.
Full experimental setting details are provided in the App.~\ref{sec:detailed_setup}.
Our analysis is organized into five parts.

\begin{table}[t!]
\centering
\vskip  -0.25in
\caption{\footnotesize Performance comparison under different rescaling strategies. 
`GA' denotes Gated Attention. 
\textbf{Attn R} and \textbf{Norm R} refer to the rescaling mechanisms in softmax attention and RMSNorm, respectively. 
`Gating' indicates learned gating-based rescaling; `Outlier-Driven' denotes outlier-driven rescaling; `Restrict' means rescaling is constrained due to activation clipping. 
`DyT' refers to the pointwise Dynamic Tanh function. 
`GLU' is a SwiGLU variant with sigmoid activation; all other configs use SwiGLU.
\textbf{LR} denotes peak learning rates.
\textbf{IDs} denotes row ID, \textbf{C IDs} denotes the compared row ID.
For \textbf{Outliers}, we retain the first two significant digits in the table. Deeper~\smash{\setlength{\fboxsep}{1pt}\colorbox{blue!10}{blue}} means larger activation magnitude; 
for \textbf{Final Loss}, deeper~\smash{\setlength{\fboxsep}{1pt}\colorbox{green!10}{green}} means lower loss. \textbf{Gap}, denotes the relative loss gap between \textbf{IDs} and \textbf{C IDs},~\smash{\setlength{\fboxsep}{1pt}\colorbox{red!10}{positive}} values indicate degradation and~\smash{\setlength{\fboxsep}{1pt}\colorbox{green!10}{negative}} values indicate improvement. 
`-' denotes divergence.}
\vskip  -0.05in
\label{tab:attn-rescale-ablation}
\resizebox{\textwidth}{!}{%
\begin{tabular}{clccccrccllc}
\toprule
\textbf{IDs} & \textbf{Basic Config} & \textbf{Additional Config} & \textbf{Attn R} & \textbf{Norm R} & \textbf{LR} & \textbf{Outliers} & \textbf{Final Loss} & \textbf{IDs} & \textbf{C IDs} & \textbf{Gap} \\
\midrule
\multicolumn{11}{c}{Full Attention} \\
\midrule
(1) & Full Attention & - & Outlier-Driven & Outlier-Driven & $4.3\times10^{-3}$ & \cellcolor{outlier7}$6,000$ & \cellcolor{loss6}1.964 & (1) &  & - \\
(2) & Full Attention & GA & Gating & Outlier-Driven & $4.3\times10^{-3}$ & \cellcolor{outlier6}$2,800$ & \cellcolor{loss6}1.957 & (2) & (1) & \cellcolor{gap_green3}-0.007 \\
\midrule
\multicolumn{11}{c}{Linear \& Hybrid Attention (Remove token mixing normalizations.)} \\
\midrule
(3) & Linear Attention & - & None & Outlier-Driven & $4.3\times10^{-3}$ & \cellcolor{outlier4}$510$ & \cellcolor{loss7}1.933 & (3) &  & - \\
(4) & Hybrid & - & Outlier-Driven & Outlier-Driven & $4.3\times10^{-3}$ & \cellcolor{outlier5}$1,800$ & \cellcolor{loss7}1.926 & (4) & (3) & \cellcolor{gap_green3}-0.007 \\
(5) & Hybrid & GA & Gating & Outlier-Driven & $4.3\times10^{-3}$ & \cellcolor{outlier5}$1,100$ & \cellcolor{loss7}1.921 & (5) & (4) & \cellcolor{gap_green2}-0.005 \\
\midrule
\multicolumn{11}{c}{Dynamic Tanh (Replace normalizations with pointwise function.)} \\
\midrule
(6) & DyT & - & None & None & $1.0\times10^{-3}$ & - & - & (6) &  & - \\
(7) &DyT & - & None & None & $5.0\times10^{-4}$ & \cellcolor{outlier2}$73$ & \cellcolor{loss1}2.216 & (7) & (1) & \cellcolor{gap_red3}+0.259 \\
(8) & DyT & GA & Gating & Outlier-Driven & $2.0\times10^{-3}$ & \cellcolor{outlier1}$32$ & \cellcolor{loss2}2.041 & (8) & (2) & \cellcolor{gap_red2}+0.084 \\
(9) & DyT & GA, GateDyT & Gating & Gating & $2.0\times10^{-3}$ & \cellcolor{outlier2}$53$ & \cellcolor{loss3}1.969 & (9) & (20) & \cellcolor{gap_red2}+0.018 \\
\midrule
\multicolumn{11}{c}{Clipping (Directly constrain outliers.)} \\
\midrule
(10) & Full Attention & clip 10 & Restrict & Restrict & $4.3\times10^{-3}$ & - & - & (10) & (1) & - \\
(11) & Full Attention & clip 100 & Restrict & Restrict & $4.3\times10^{-3}$ & - & - & (11) & (1) & - \\
(12) & Full Attention & clip 1000 & Restrict & Restrict & $4.3\times10^{-3}$ & \cellcolor{outlier5}$1,000$ & \cellcolor{loss5}1.970 & (12) & (1) & \cellcolor{gap_red1}+0.006 \\
(13) & Full Attention & GA, clip 10 & Restrict & Restrict & $4.3\times10^{-3}$ & \cellcolor{outlier1}$10$ & \cellcolor{loss5}1.960 & (13) & (2) & \cellcolor{gap_red1}+0.003 \\
(14) & Full Attention & GA, clip 1000 & Restrict & Restrict & $4.3\times10^{-3}$ & \cellcolor{outlier5}$1,000$ & \cellcolor{loss5}1.958 & (14) & (2) & \cellcolor{gap_red1}+0.001 \\
\midrule
\multicolumn{11}{c}{GLU Variants (Constrain outliers through architecture modifications.)} \\
\midrule
(15) & Full Attention & GLU & Restrict & Restrict & $4.3\times10^{-3}$ & \cellcolor{outlier5}$1,300$ & \cellcolor{loss5}1.975 & (15) & (1) & \cellcolor{gap_red1}+0.011 \\
(16) & Full Attention & GA, GLU & Gating & Restrict & $4.3\times10^{-3}$ & \cellcolor{outlier4}$800$ & \cellcolor{loss6}1.955 & (16) & (2) & \cellcolor{gap_green2}-0.002 \\
(17) & Full Attention & GA, PreAffine, GLU & Gating & Outlier-Driven & $4.3\times10^{-3}$ & \cellcolor{outlier2}$150$ & \cellcolor{loss6}1.952 & (17) & (19) & \cellcolor{gap_green2}-0.002 \\
(18) & Full Attention & GA, GatedNorm, GLU & Gating & Gating & $4.3\times10^{-3}$ & \cellcolor{outlier3}$280$ & \cellcolor{loss6}1.948 & (18) & (20) & \cellcolor{gap_green2}-0.003 \\
\midrule
\multicolumn{11}{c}{PreAffine \& GatedNorm (Residual sink reduction methods.)} \\
\midrule
(19) & Full Attention & GA, PreAffine & Gating & Outlier-Driven & $4.3\times10^{-3}$ & \cellcolor{outlier4}$640$ & \cellcolor{loss6}1.954 & (19) & (2) & \cellcolor{gap_green2}-0.003 \\
(20) & Full Attention & GA, GatedNorm & Gating & Gating & $4.3\times10^{-3}$ & \cellcolor{outlier4}$430$ & \cellcolor{loss6}1.951 & (20) & (2) & \cellcolor{gap_green3}-0.006 \\
(21) & Linear Attention & GatedNorm & None & Gating & $4.3\times10^{-3}$ & \cellcolor{outlier2}$110$ & \cellcolor{loss7}1.929 & (21) & (3) & \cellcolor{gap_green2}-0.004 \\
(22) & Hybrid Attention & GA, GatedNorm & Gating & Gating & $4.3\times10^{-3}$ & \cellcolor{outlier4}$780$ & \cellcolor{loss7}1.918 & (22) & (5) & \cellcolor{gap_green2}-0.003 \\
\bottomrule
\end{tabular}%
}
\vskip  -0.1in
\end{table}

In Sec.~\ref{sec:dyt}, we show that replacing normalization layers with point-wise functions such as Dynamic Tanh (DyT)~\citep{zhu2025transformers,chen2025stronger} significantly reduces outliers. 
However, as DyT cannot provide outlier-driven rescaling, both training stability and final performance degrade.  

In Sec.~\ref{sec:clip}, we demonstrate that even when normalization is retained, directly constraining outliers (via activation clipping) breaks the outlier-driven rescaling mechanism and harms model performance, sometimes causing training divergence. 
This also explains why architectural changes that suppress outlier generation, such as using sigmoid-based GLU variants, tend to underperform~\citep{shazeer2020glu}.  

In Sec.~\ref{sec:pre_affine}, we show that outliers can be losslessly transferred from activations into learnable parameters: by introducing a lightweight learnable vector before normalization, the model can still perform outlier-driven rescaling without requiring large values in the residual stream, but use the amplified projections after the learnable vector.  

In Sec.~\ref{sec:gate}, we show that enabling rescaling via gating reduces residual sinks without performance degradation.

In Sec.~\ref{sec:arch_choice}, we find that once gating-based rescaling is introduced and the model’s reliance on outliers is reduced, its sensitivity to architectural choices diminishes. 
Specifically, DyT achieves stable convergence, and sigmoid-based GLU matches or even surpasses SwiGLU regarding performance.

\subsection{Removing Normalizations Reduces Outliers with Degraded Stability and Performance}
\label{sec:dyt}

Normalizations in transformer layers are in two places: softmax in attention and normalization layers (e.g., RMSNorm).
In attention, prior work shows that softmax normalization is a primary cause of attention sinks~\citep{gu2024attention,an2025systematic}. 
When softmax is replaced with sigmoid attention~\citep{gu2024attention}, or when the denominator is combined with a learnable bias~\citep{sun2024massive, dong2024hymba, agarwal2025gpt}, attention sinks disappear.  

Existing work finds that tokens exhibiting attention sinks produce value vectors with smaller norms than other tokens~\citep{sun2024massive,an2025systematic}.
From the perspective of outlier-driven rescaling, the presence of the attention sink allows the model to adjust the relative contribution of near-zero contributions (from sink tokens) versus others in the attention output, controlling the norm of the attention result. 
This interpretation also explains the training instability observed in sigmoid attention: without the normalization-induced rescaling, initial attention outputs have large norms~\citep{ramapuram2024theory}. 

Our experiments find that replacing softmax-based attention with linear attention~\citep{yang2024gated} (without normalization in the token-mixing step) also reduces MA.
As shown in rows (3)--(5) of Tab.~\ref{tab:attn-rescale-ablation}, the peak activation drops to 510 for the linear attention model and 1100 for a hybrid model combining full and linear attention in a 1:3 ratio, compared to 6000 for the full attention baseline. Notably, while linear attention eliminates MA, residual sinks persist.

For normalization layers in the residual, prior work shows that replacing RMSNorm with DyT---defined as \( \text{DyT}(x) = \gamma \cdot \tanh(\alpha x) + \beta \)---significantly reduces outliers~\citep{he2024understanding,owen2025refined}. 
The key difference lies in rescaling: in RMSNorm, each dimension is scaled based on statistics of the entire hidden state (e.g., root-mean-square), allowing the outlier in one dimension to influence all other dimensions. 
In contrast, DyT uses only pointwise operations, so no dimension directly influences another.

Our experiments confirm this limitation. 
When training DyT models with the same learning rate as the baseline, optimization diverges rapidly. 
To address this, we conduct hyperparameter sweeps for all DyT inclusive settings, evaluating learning rates in \(\{5 \times 10^{-4}, 1 \times 10^{-3}, 2 \times 10^{-3}, 4 \times 10^{-3}\}\) and reporting the best performing configuration. 
The original DyT variant only converges at the smallest learning rate \(5 \times 10^{-4}\), yielding a peak activation magnitude of 73 but suffering a significant performance drop of +0.259 in loss compared to the baseline (Tab~\ref{tab:attn-rescale-ablation} row (7) versus row (1)). 
This suggests that the outlier-driven rescaling effect is essential for both training stability and final performance. 

Taken together, these results show that removing normalization breaks the outlier-driven rescaling mechanism. 
Consequently, the model stops generating outliers, but usually at the cost of degraded training stability and performance.

\subsection{Directly Clipping or Constraining Outliers Hurts Stability and Performance}
\label{sec:clip}

We now examine the effect of suppressing outliers while preserving normalization. 
To isolate the impact of different outliers, we consider two settings:  
(1) models exhibiting attention sinks, MA, and residual sinks;  
(2) models where attention sinks and MA are reduced via Gated Attention (GA)~\citep{bondarenko2023quantizable,an2025systematic,qiu2025gated}, exhibiting primarily residual sinks.

First, we apply activation clipping to the residual of the full attention baseline (Tab.~\ref{tab:attn-rescale-ablation}, row (1)), capping activations above a threshold at that value. 
When the clipping threshold is set to 100 or lower, training diverges early. 
At a threshold of 1000, the loss curve shows frequent spikes and converges to a higher final loss (+0.006, row (12)). 
This aligns with prior observations that clipping MA or attention sinks in the models after training severely degrades performance, often yielding near-random outputs.

Second, we combine GA, which reduces attention sinks and MA, with residual clipping. 
We find that once explicit rescaling is reintroduced via GA in the softmax attention, even aggressive clipping (at 10) permits convergence but still harms performance (+0.003, row (13)). 
Moreover, under the same setting, clipping at 1000 incurs a much smaller performance drop (+0.001, row (14)).

These results imply two key points:  
\textbf{(i)} The outlier-driven rescaling mechanism in attention is the primary source of instability when disrupted;  
\textbf{(ii)} Residual sinks also contribute meaningfully to performance, and directly constraining them without compensation leads to degradation.

Beyond clipping, we explore a less intrusive method to limit outliers by modifying a seemingly unrelated component, the activation function, thereby further validating the above conclusions. 
Prior work observes that outliers in transformers predominantly originate from FFN~\citep{oh2024house, yona2025interpreting}. Most modern models employ Gated Linear Units variants, particularly SwiGLU~\citep{shazeer2020glu}:  
\[
\text{SwiGLU}(\mathbf{x}) = \mathbf{x}_{\text{down}} \big( (\mathbf{W}_{\text{up}} \mathbf{x}) \odot \text{swish}(\mathbf{W}_{\text{gate}} \mathbf{x}) \big).
\]  
In SwiGLU, outliers emerge in the terms \(\mathbf{W}_{\text{up}} \mathbf{x} \odot\text{swish}(\mathbf{W}_{\text{gate}} \mathbf{x})\), and are amplified by \(\mathbf{W}_{\text{down}}\).

Conceptually, replacing the swish activation with sigmoid—which has a bounded range of \((0, 1)\)—naturally constrains the magnitude of FFN outputs and effectively suppresses outlier generation. This modification reduces SwiGLU to standard GLU. As shown in Tab.~\ref{tab:attn-rescale-ablation}, GLU exhibits significantly smaller outlier magnitudes at convergence compared to SwiGLU (1300 vs. 6000), but incurs a performance drop of +0.011. 
This aligns with prior findings that GLU variants using sigmoid underperform those using swish or GELU~\citep{shazeer2020glu}.
In Sec.~\ref{sec:arch_choice}, we further observe that when the model’s reliance on outlier-driven rescaling is reduced through explicit rescaling like gating, \textit{GLU can even slightly outperform SwiGLU}.

\subsection{Fusing Outliers into Parameters}
\label{sec:pre_affine}

The previous sections establish that outliers serve a functional role in normalizations, and removing them without compensation harms performance.
Theoretically, we proof the upper bound on the feature norm after RMSNorm decreases as the outlier increases (App.~\ref{proof}), allowing outliers to rescale the feature norm.
We now ask: \textit{can we preserve this functionality while reducing explicit outliers?}

\begin{figure}[!t]
    \vskip  -0.3in
    \centering
    \includegraphics[width=\linewidth]{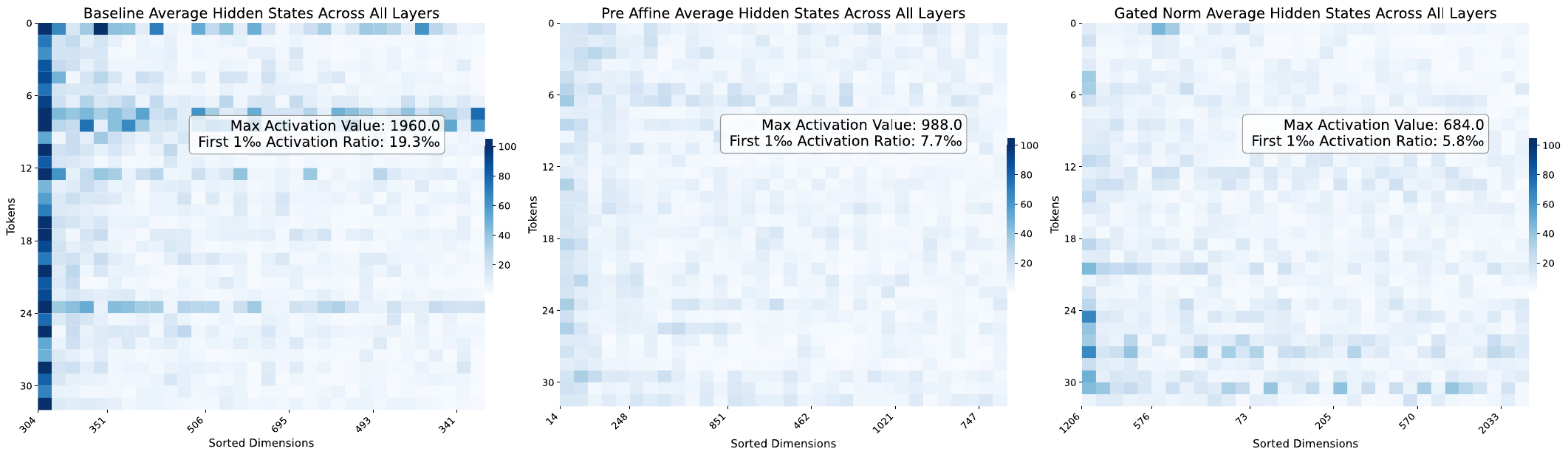}
    \vskip  -0.05in
    \caption{Reduced residual sinks. For the same input, the baseline (left) has dimension 304 consistently produces large activations across all tokens. 
    This phenomenon is substantially mitigated in both the PreAffine (middle) and GatedNorm (right) variants.}
    \label{fig:reduce_sink}
    \vskip  -0.2in
\end{figure}

Prior work shows that learnable sink tokens can shift attention sinks from input logits to fixed parameters~\citep{sun2024massive, dong2024hymba, agarwal2025gpt}, removing outliers from real tokens while preserving the outlier-driven rescaling. 
Inspired by this, we introduce a learnable element-wise scaling vector (termed \textit{PreAffine}) before RMSNorm. Specifically:

\[
\text{PreAffineRMSNorm}(\mathbf{x}) = \text{RMSNorm}(\lambda_1 \odot \mathbf{x}),
\]

where \(\lambda_1 \in \mathbb{R}^d\) is a trainable vector.
Since residual sinks appear consistently in the same dimensions across nearly all tokens, \(\lambda_1\) can learn to amplify those specific dimensions. 
Thus, even if the input activation \(\mathbf{x}\) contains no outliers, the scaled input \(\lambda_1 \odot \mathbf{x}\) can still contain large values in selected dimensions, enabling outlier-driven rescaling as usual.

After adding PreAffine, the maximum activation in the network decreases from 2800 to 640, and the final loss slightly improves (-0.003, Tab.~\ref{tab:attn-rescale-ablation}, row (19)).  
We also evaluate various outlier reduction methods on a 24.6B-A1.7B hybrid MoE model (configuration detailed in Section~\ref{sec:scaling}) and report the corresponding activation statistics in Fig.~\ref{fig:reduce_sink}.
The left panel shows the baseline with GA suppressing MA; here, dimension 304 consistently exhibits higher activation than other dimensions, indicating a residual sink. 
In the middle panel, we apply PreAffine to the same model. The persistent high activation in any single dimension disappears, and the peak activation drops from 1960 to 988. 

Importantly, the rescaling capability of \(\lambda_1\) differs from that of the standard RMSNorm parameter \(\lambda\). 
\(\lambda_1\) interacts with RMSNorm: large values in a few dimensions of \(\lambda_1\) control the RMS of the scaled input, thereby rescaling the strength of non-outlier dimensions.
We analyze the parameters \(\lambda\) (the standard RMSNorm weight) and \(\lambda_1\) (the PreAffine) across models; full results are provided in App.~\ref{app:rmsnorm_weight}. 
We identify the dimensions with the largest deviations from 1, as these affine parameters exert the strongest influence. Our observations are as follows:

(1) In the baseline model, most dimensions of \(\lambda\) remain close to 1. However, dimension 304 consistently deviates from 1 across all layers, reaching a minimum value of 0.004. 
Notably, the residual sink in the baseline is also in dimension 304. 
This indicates that the large activation caused by the residual sink is immediately scaled down after normalization. 
This suggests that once a dimension fulfills its role in outlier-driven rescaling, its downstream influence is intentionally dampened. 
This mirrors the observation that the attention sink token's value vectors exhibit smaller norms.

(2) In the model equipped with PreAffine, the deviation of \(\lambda_1\) from 1 is significantly larger than that of \(\lambda\). 
For example, \(\lambda_1\) in dimension 1326 reaches 7.19, while the corresponding \(\lambda\) in the same dimension is only 0.06. 
This further supports the view that outliers are used to shape representations with normalization, and their direct contribution is suppressed.

\subsection{GatedNorm: Explicitly Enabling Rescaling}
\label{sec:gate}

Although PreAffine reduces outliers in the residual stream, outliers still appear within the normalization computation (i.e., in \(\lambda_1 \odot \mathbf{x}\)). 
To address this, we draw inspiration from GA and introduce GatedNorm: an element-wise low-rank self-gating mechanism applied after every normalization layer.
Formally, given \(\mathbf{y} = \text{RMSNorm}(\mathbf{x})\), we compute:
\[
\mathbf{y}_g = \sigma\big(\mathbf{W}_{\text{up}}(\text{swish}(\mathbf{W}_{\text{down}}(\mathbf{y})))\big), \quad \mathbf{y}' = \mathbf{y}_g \odot \mathbf{y},
\]
where \(\mathbf{W}_{\text{down}} \in \mathbb{R}^{d \times r}\), \(\mathbf{W}_{\text{up}} \in \mathbb{R}^{r \times d}\), \(r \ll d\) (e.g., \(r = 16\)), \(\sigma\) is sigmoid activation.

Notably, GatedNorm adds only 3.7M parameters, which is approximately 2\% of the total in a 2B model. 
To maintain parameter parity, we slightly reduce FFN capacity. 
In the 2B dense model GatedNorm incurs about 5\% latency overhead, and this overhead further decreases as model size increases, especially in MoEs. 
More detailed performance analysis is provided in the App.~\ref{app:efficiency}.
We examine GatedNorm on top of several settings in Tab.~\ref{tab:attn-rescale-ablation} that still exhibit residual sinks. 
As shown in rows (20)–(22), GatedNorm further reduces both loss and outlier magnitude of models with GatedAttention (including full attention, hybrid attention, and linear attention).

Figure~\ref{fig:reduce_sink} (right) shows that, on a 24.6B-A1.7B hybrid MoE model, GatedNorm also suppresses residual sinks. 
We analyze the learned scaling parameters \(\lambda\) in models using GatedNorm (App.~\ref{app:rmsnorm_weight}, Fig.~\ref{fig:rmsnorm_gated}). 
The maximum deviation of \(\lambda\) from 1 is only 0.73, compared with 0.004 in the baseline and 0.06 in the PreAffine model, indicating that when the network no longer relies on outlier-driven rescaling, the need to suppress any particular dimension after normalization disappears. 
Consequently, normalization outputs become smoother and quantization-friendly.

We also compare different gating variants, focusing on two design choices: the gating granularity (elementwise (score shape $d$) versus tensorwise (score shape $1$)) and the activation function (sigmoid, tanh, silu, identity). Our findings are as follows.

First, elementwise gating with a sigmoid activation yields the most significant performance improvement over the baseline. 
Tensorwise gating with sigmoid reduces residual sinks to a similar extent as elementwise gating, but its final performance is close to the baseline and consistently inferior to the elementwise variant. 
This suggests that while both granularities can mitigate outliers, which supports the outlier-driven rescaling hypothesis, finer-grained (elementwise) rescaling enables more effective modulation and thus better performance.

Second, when using elementwise gating, replacing sigmoid with tanh, SiLU, or no activation (similar to adaLN~\citep{perez2018film,xu2019understanding,peebles2023scalable,karras2024analyzing} in DiT) leads to unstable outlier dynamics during training. 
This implies that the bounded nature of sigmoid and its fine-grained control near zero are beneficial for stable rescaling, consistent with~\citet{chen2025stronger}. Moreover, under tensorwise gating, all non-sigmoid activations, including tanh, SiLU, or none, cause training divergence, further highlighting the necessity of a well-behaved, bounded activation like sigmoid for stable gating.

\subsection{GatedNorm Improve Robustness to Architecture Choice}
\label{sec:arch_choice}

Sec.~\ref{sec:dyt} and~\ref{sec:clip} show that both DyT and sigmoid-based GLU underperform the baseline. 
One possible explanation is their architectures inherently restrict the outlier-driven rescaling mechanism. 
In this section, we investigate whether explicitly providing rescaling, via GatedNorm, can recover their performance by reducing reliance on outliers.

\begin{wrapfigure}{r}{0.55\linewidth}
    \vspace{-0.15in}
    \centering
    \includegraphics[width=\linewidth]{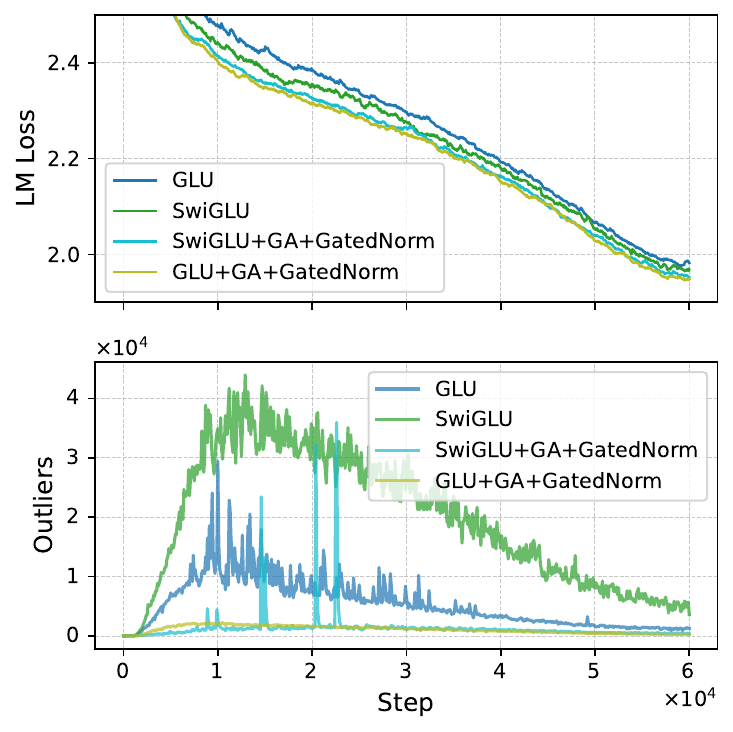}
    \vspace{-0.2in}
    \caption{\footnotesize SwiGLU and GLU with different rescaling method.}
    \label{fig:swiglu}
    \vspace{-0.15in}
\end{wrapfigure}

We first equip DyT with GA (row 8 in Tab.~\ref{tab:attn-rescale-ablation}) to provide explicit rescaling in the attention module. 
This enables the model to train stably at the baseline’s learning rate (4.3e-3), though its optimal learning is 2e-3. 
This further confirms the critical role of attention rescaling in training stability.
We then apply a low-rank self-gating mechanism after the DyT layer, analogous to GatedNorm, resulting in GatedDyT (row 9  in Tab.~\ref{tab:attn-rescale-ablation}). 
With this addition, the performance gap between DyT and the RMSNorm baseline narrows from 0.084 to 0.018, highlighting the importance of explicit rescaling not only in attention but also at the normalization layer itself.

We further compare SwiGLU and GLU before and after incorporating GatedNorm. 
As shown in Fig.~\ref{fig:swiglu} (bottom), vanilla SwiGLU generates maximum activations exceeding \(4 \times 10^4\) during training, while vanilla GLU peaks around \(1 \times 10^4\). Lacking outlier-driven rescaling, GLU converges to a higher loss (+0.011).
After employing GatedNorm, GLU slightly outperforms SwiGLU under the same setup (-0.002 in row 16; -0.003 in row 17). Fig.~\ref{fig:swiglu} (top) shows that, with gating, the GLU loss curve improves from the worst (blue) to the best (green) among all variants.

This suggests that the performance gap between GLU and SwiGLU primarily stems from their differing capacities to support outlier-driven rescaling. 
This also explains why higher-order activation functions that more readily produce outliers, such as \(\text{ReLU}^2\)~\citep{zhang2024relu} or PolyNorm~\citep{zhuo2024polynomial}, can be advantageous when such rescaling is required. 
When rescaling is explicitly provided via GatedNorm, the model becomes robust to architectural choices that affect outlier generation.

\begin{table}[!t]
\centering
\vskip  -0.05in
\caption{\footnotesize Performance of different configurations. 
`GA' denotes Gated Attention. 
`MMLU-R' denotes MMLU-Redux; `MMLU-P' denotes MMLU-Pro; `GPQA-D' denotes GPQA-Diamond; `S-GPQA' denotes SuperGPQA. 
`W4A4' indicates 4-bit weight and activation quantization; `SQ' denotes smooth quantization.
All models are based on the 7B-A2B or 24B-A3B architecture unless otherwise noted.}
\vskip  -0.05in
\label{tab:quantization-ablation}
\resizebox{\textwidth}{!}{
\begin{tabular}{l l l ccc ccc ccc cc c}
\toprule
 &  &  &
\multicolumn{3}{c}{\textbf{Knowledge}} &
\multicolumn{3}{c}{\textbf{STEM}} &
\multicolumn{3}{c}{\textbf{Code}} &
\multicolumn{2}{c}{\textbf{Multilingual}} &
 \\
\cmidrule(lr){4-6} \cmidrule(lr){7-9} \cmidrule(lr){10-12} \cmidrule(lr){13-14}
 & \textbf{Type} & \textbf{Add Config} &
MMLU-R & MMLU-P & S-GPQA &
GPQA-D & GSM8k & Math &
Crux & MultiPL\_E & MBPP &
MMMLU & MGSM & \textbf{Avg} \\
\midrule
\multicolumn{15}{c}{MoE-7B-A2B, 1.2T tokens} \\
\midrule
(1) & BF16 & GA & 59.59 & 31.91 & 16.39 & 26.93 & 66.22 & 45.36 & 44.38 & 34.63 & 50.04 & 46.16 & 37.40 & 41.73 \\
(2) & BF16 & GA, PreAffine & 61.59 & 32.00 & 18.75 & 27.40 & 67.10 & 44.26 & \textbf{46.75} & \textbf{37.99} & 49.20 & \textbf{46.42} & \textbf{39.90} & 42.85 \\
(3) & BF16 & GA, GatedNorm & \textbf{61.71} & \textbf{33.46} & \textbf{19.05} & \textbf{29.90} & \textbf{68.04} & \textbf{46.66} & 45.56 & 34.81 & \textbf{51.00} & 45.76 & 38.27 & \textbf{43.11} \\
\midrule
\multicolumn{15}{c}{24B-A3B, 500B tokens} \\
\midrule
(4) & BF16 & GA & 67.49 & 46.02 & 25.64 & 31.66 & 79.45 & 53.92 & 58.00 & 45.88 & 56.80 & \textbf{55.67} & 59.27 & 52.71 \\
(5) & BF16 & GA, PreAffine & 67.96 & \textbf{48.48} & 24.34 & 32.64 & 82.07 & 52.62 & 59.13 & 46.90 & 54.80 & 55.46 & 58.46 & 52.99 \\
(6) & BF16 & GA, GatedNorm & \textbf{69.70} & 47.13 & \textbf{25.81} & \textbf{33.87} & \textbf{82.26} & \textbf{62.70} & \textbf{60.12} & \textbf{47.52} & \textbf{58.40} & 55.47 & \textbf{59.72} & \textbf{54.79} \\
\midrule
\multicolumn{15}{c}{24B-A3B, 500B tokens, FP4 Quantization} \\
\midrule
(7) & W4A4 & GA & 65.77 & 45.71 & 24.73 & 31.98 & 76.38 & 52.66 & 56.50 & 44.41 & 56.60 & 53.21 & 51.85 & 50.89 \\
(8) & W4A4+SQ & GA & 66.63 & 44.86 & 24.68 & 32.77 & 77.45 & 52.82 & 55.19 & 46.83 & 56.60 & 53.20 & 52.55 & 51.23 \\
(9) & W4A4 & GA, PreAffine & 66.05 & 43.61 & 24.14 & 31.60 & 78.70 & 49.50 & 56.69 & 44.01 & 53.00 & 51.35 & 49.58 & 49.84 \\
(10) & W4A4+SQ & GA, PreAffine & 67.17 & 43.44 & 23.77 & 31.66 & 79.42 & 49.05 & 57.56 & 42.76 & 55.00 & 51.98 & 50.72 & 50.23 \\
(11) & W4A4 & GA, GatedNorm & 66.92 & 46.36 & \textbf{25.59} & 32.70 & \textbf{81.35} & 59.40 & \textbf{59.13} & 46.97 & \textbf{58.80} & \textbf{53.70} & \textbf{56.78} & 53.43 \\
(12) & W4A4+SQ & GA, GatedNorm & \textbf{67.35} & \textbf{47.15} & 25.11 & \textbf{33.62} & 80.59 & \textbf{61.88} & 58.06 & \textbf{47.18} & 58.80 & 53.39 & 56.00 & \textbf{53.56} \\
\bottomrule
\end{tabular}
}
\vskip  -0.15in
\end{table}

\section{Scaling Outlier Mitigations and Deployment-Level Quantization}
\label{sec:scaling}

In this section, we evaluate different combinations of GatedAttention and GatedNorm, or PreAffine, in large-scale settings. 
As Tab.~\ref{tab:attn-rescale-ablation} row (5) shows, hybrid models exhibit advantages, we conduct experiments on efficient hybrid MoE models following Qwen3-Next, on two settings:  
(1) a 7.4B-parameter model with 1.7B activated parameters (MoE-7B-A-2B), trained on 1.2T tokens;  
(2) a 24.6B-parameter model with 2.7B activated parameters (MoE-24B-A3B), trained on 500B tokens.  
In this setting, GatedNorm incurs less than 3\% latency overhead.
Full details are provided in App.~\ref{app:scaling_set}.

\begin{wrapfigure}{r}{0.55\linewidth}
    \centering
    \vskip  -0.2in
    \includegraphics[width=\linewidth]{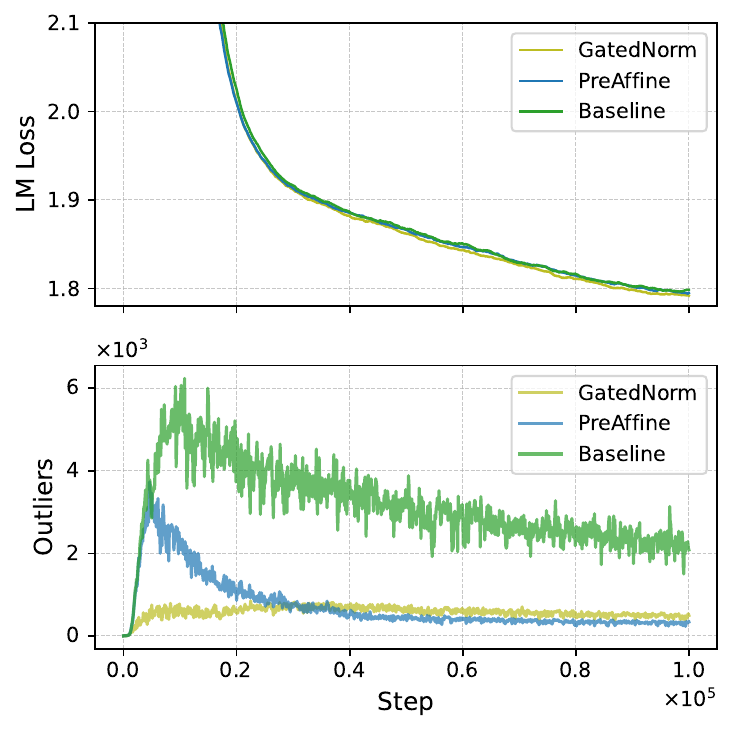}
    \vskip  -0.1in
    \caption{\footnotesize Loss and outliers of MoE-24B-A3B models.}
    \label{fig:turbo_loss}
    \vskip  -0.2in
\end{wrapfigure}

Fig~\ref{fig:turbo_loss} shows the training loss curves and outliers for the MoE-24B-A3B model. 
Our observations include:  
(1) Baseline's outliers rise rapidly in early training and gradually decay as the learning rate decreases. 
The PreAffine model also exhibits an initial outlier surge (within the first 10\% of training), but quickly diminishes thereafter.
One possible explanation is that the learnable scaler \(\lambda_1\) initially lacks sufficient magnitude to fully support outlier-driven rescaling, forcing the model to rely temporarily on activation outliers; as training progresses and specific dimensions of \(\lambda_1\) grow large, outliers are `absorbed' into the parameters. 
GatedNorm shows no early surge of outliers and maintains consistently low activation magnitudes throughout training, indicating reduced dependence on outlier-driven rescaling.  
(2) A clear performance gap emerges in the mid-to-late stages, with GatedNorm achieving a lower final loss.

Tab.~\ref{tab:quantization-ablation} compares different outlier mitigation strategies across BF16 and quantized settings.
Detailed quantization configs are in App.~\ref{app:quant}
Since FP8 quantization yields relatively minor changes, we focus on aggressive FP4 W4A4 quantization. 
Results show consistent gains from PreAffine and GatedNorm across both model sizes, with GatedNorm delivering larger improvements. 
Notably, GatedNorm achieves +1.0 point gain over the baseline on knowledge tasks and exceeds +2.0 points on STEM and Code tasks.

Under FP4 quantization, we observe:  
(1) for models using only GA and PreAffine, SmoothQuant~\citep{xiao2023smoothquant} provides an average +0.5 point gain, greater than the gain for GatedNorm, suggesting that GatedNorm’s rescaling largely mitigates outlier sensitivity;  
(2) GatedNorm demonstrates the smallest performance drop after FP4 quantization (-1.23 points), compared to GA (-1.50) and PreAffine (-2.76). 
On the MGSM benchmark, only GatedNorm maintains degradation within 5 points; all other methods incur nearly 10-point losses.

This superior quantization robustness aligns with our observation of an property of gating-based rescaling: it exhibits outlier-suppressing behavior, where dimensions with large \(|y|\) tend to receive smaller gating scores \(y_g\), yielding smoother final activations. 
Overall, PreAffine relocates outliers, similar in spirit to learnable sink and SmoothQuant~\citep{xiao2023smoothquant}, but still relies on them to perform outlier-driven rescaling. 
In contrast, GatedNorm provides explicit gating, producing inherently smoother activations throughout the network and achieving superior quantization robustness.

\section{Related Works}

Outliers in transformers have been widely studied. 
In BERT models, outliers in fixed-dimensions are primarily attributed to the weight and bias parameters in LayerNorm~\citep{bondarenko2021understanding,kovaleva2021bert,wei2022outlier}, and are closely associated with the attention patterns of special tokens~\citep{puccetti2022outlier}. 
This phenomenon resembles the attention sink observed in GPT-style models and significantly impacts model performance~\citep{kovaleva2021bert,xiao2023efficient}. 
The outliers discussed in BERT largely correspond to MA~\citep{sun2024massive,gu2024attention, yu2024unveiling} in autoregressive models. 
These MA typically originate in semantically sparse special tokens, emerge early in FFNs, and propagate through the residual stream, continuously influencing attention distributions in subsequent layers~\citep{sun2024massive,oh2024house,gu2024attention,yona2025interpreting}.

Several works identify input-independent outliers that consistently appear in fixed dimensions of GPT models~\citep{dettmers2022gpt3}. 
These outliers are not directly related to attention sinks~\citep{he2024understanding,an2025systematic}. 
\citet{he2024understanding} further attributes these outliers to the normalizations themselves, showing that they persist even when the LayerNorm weights are removed.
As outliers hurt both training and inference quantization, a number of approaches aim to mitigate their impact. 
Common techniques include row-wise, channel-wise, group-wise scaling to limit the quantization error caused by outliers~\citep{yao2022zeroquant,xiao2023smoothquant,wei2023outlier,nvfp4}, as well as Hadamard transformations to redistribute outlier across dimensions~\citep{xi2023training, wang2025bitnet}.

Another studies focus on reducing outliers during training. 
From the \textit{optimization} perspective, strategies such as increasing weight decay, gradient clipping~\citep{ahmadian2023intriguing}, constraining weight variance~\citep{owen2025variance,owen2025refined,xie2026controlled}, or adding explicit regularization loss terms~\citep{liang2025tweo} can suppress outliers. 
Some studies also examine whether the Adam optimizer causes the outliers in pre-training~\citep{kaul2024attention, he2024understanding, xie2026controlled}. 
From the \textit{architectural} viewpoint, prior work has noted a strong connection between outliers and normalization, and proposed removing normalization to eliminate outliers~\citep{he2024understanding,owen2025refined,owen2025variance}. 
Our work shows that, through architectural interventions that explicitly replace outlier-driven rescaling, models can be trained with much smaller activation magnitudes while still using Adam, standard training recipes and normalizations.

Most closely related to our work are studies that investigate the functional role of outliers. 
\citet{bondarenko2023quantizable, an2025systematic} propose that attention outliers act as context-aware scaling factors, and demonstrate that introducing gating-based scaling in the attention reduces attention outliers. 
\citet{karras2020analyzing} identify normalization as the source of outliers in intermediate feature maps of StyleGAN, where these outliers serve to scale signals during normalization. 
It further shows that input-dependent convolutional weights generated via gating can replicate this scaling effect even without normalization.
This gating-based scaling is also adopted in adaLN~\citep{perez2018film,xu2019understanding,peebles2023scalable,karras2024analyzing}. 
Our work extends this insight to residual sinks in LLMs, highlighting the widespread presence of outlier-driven rescaling across transformers and providing systematic evidence through a series of targeted architectural interventions.

\section{Conclusion}

This paper argues that outliers in LLMs are not mere artifacts but have functional roles.
They work together with normalization mechanisms (softmax and RMSNorm) to perform outlier-driven rescaling, which rescales the magnitude of non-outlier features. 
This mechanism is essential for stable training and strong performance. 
Removing outliers and breaking outlier-driven rescaling harms the model.
By explicitly providing gating-based rescaling, we can reduce activation outliers while maintaining or even improving performance.
Moreover, explicitly enabling rescaling reduces sensitivity to
architecture choice.
These approaches also yield smoother activations and significantly better quantization robustness, especially under aggressive low-bit settings.

\section*{Limitations}

This work empirically demonstrates the importance of outlier-driven rescaling in network training and shows that models can leverage normalization, such as RMSNorm, to adjust feature norms. 
However, we do not investigate why such rescaling is necessary for effective training or representation learning. A deeper theoretical understanding of the role of rescaling remains an open question.

\bibliography{custom}
\bibliographystyle{colm2024_conference}

\appendix

\section{Appendix}

\subsection{A Brief Calculation On How Outlier Interacts With the Normalization Layer}
\label{proof}

We will illustrate below on how residual sinks can rescale the norm of the features after the LayerNorm transformation. 
Denote the input feature as $\mathbf{h} \in \mathbb{R}^D$. 
Denote the rescaling parameters of the $\mathrm{LN}$ as $\lambda$ and suppose there is only one outlier dimension $d$.

As we observe the outlier corresponds to very small affine parameter, assume $|\lambda_d| \le \epsilon \lVert \lambda \rVert_{\infty}$. Further assume outlier corresponds to $r$ ratio of the norm of the features, that is $r = |\mathbf{h}_d| / \lVert \mathbf{h} \rVert_2$. We then have the following inequality: 
\begin{align}
\lVert \mathrm{LN}(\mathbf{h}) \rVert_{\mathrm{rms}}
\le \,\lVert \lambda \rVert_\infty\,\sqrt{(1-r^2)+\epsilon^2 r^2}.
\end{align}

This shows that the upper bound on the feature norm after LayerNorm decreases as the outlier becomes larger, allowing the network to rescale the feature norm by changing the magnitude of outliers.

To prove this inequality, let
\[
\mathbf{u} := \frac{\mathbf{h}}{\lVert \mathbf{h} \rVert_{\mathrm{rms}}},\qquad \lVert \mathbf{u} \rVert_{\mathrm{rms}}=1,\qquad |\mathbf{u}_d|=r,\qquad \sum_{i\neq d}\mathbf{u}_i^2 = 1-r^2.
\]
Then
\begin{align}
\mathrm{LN}(\mathbf{h}) &= \frac{\lambda\odot \mathbf{h}}{\lVert \mathbf{h} \rVert_{\mathrm{rms}}},\\
\lVert \mathrm{LN}(\mathbf{h}) \rVert_2 &= \frac{\lVert \lambda\odot \mathbf{h} \rVert_2}{\lVert \mathbf{h} \rVert_{\mathrm{rms}}}
= \sqrt{D}\,\frac{\lVert \lambda\odot \mathbf{h} \rVert_2}{\lVert \mathbf{h} \rVert_2}
= \sqrt{D}\,\lVert \lambda\odot \mathbf{u} \rVert_2,
\end{align}
and
\begin{align}
\lVert \lambda\odot \mathbf{u} \rVert_2^2
= \lambda_d^2 r^2 + \sum_{i\neq d}\lambda_i^2 \mathbf{u}_i^2.
\end{align}

Using $\sum_{i\neq d}\lambda_i^2 \mathbf{u}_i^2 \le \lVert \lambda_{-d} \rVert_\infty^2\sum_{i\neq d}\mathbf{u}_i^2
= \lVert \lambda_{-d} \rVert_\infty^2(1-r^2)$, we get
\begin{align}
\lVert \mathrm{LN}(\mathbf{h}) \rVert_2
\le \sqrt{D}\,\sqrt{\lVert \lambda_{-d} \rVert_\infty^2(1-r^2)+\lambda_d^2 r^2}.
\end{align}

\subsection{Comparison Between Different Outliers}
\label{app:comparision}

\begin{table*}[ht]
\centering
\caption{Comparison of attention sinks and residual sinks under the outlier-driven rescaling perspective.}
\label{tab:sink_comparison}
\resizebox{\textwidth}{!}{
\begin{tabular}{lcc}
\toprule
\textbf{Aspect} & \textbf{Attention Sink} & \textbf{Residual Sink} \\
\midrule
Occurrence & Special tokens (e.g., first token) & Most tokens, fixed dimensions \\
Associated Normalization & Softmax in attention & RMSNorm \\
Functional Role & Rescales attention output norm~\citep{an2025systematic} & Rescales RMSNorm output norm \\
Effect of Removal & Collapse to randomness~\citep{xiao2023efficient,sun2024massive} & Leads to performance degradation (Sec.~\ref{sec:clip}) \\
Downstream Suppression & Corresponding value vectors are small~\citep{sun2024massive,an2025systematic} & Corresponding RMSNorm affine weights are small (Sec.~\ref{sec:pre_affine}) \\
Reduction & Diminishes with sigmoid~\citep{ramapuram2024theory} or linear attention (Sec.~\ref{sec:dyt}) & Diminishes with pointwise functions (Sec.~\ref{sec:dyt}) \\
Absorption into Parameters & Learnable sink tokens/biases~\citep{sun2024massive} & PreAffine (Sec.~\ref{sec:pre_affine}) \\
Explicit Rescaling Alternative & Gated Attention~\citep{bondarenko2023quantizable,an2025systematic,qiu2025gated} & GatedNorm (Sec.~\ref{sec:gate}) \\
\bottomrule
\end{tabular}
}
\end{table*}

Tab.~\ref{tab:sink_comparison} summarizes the parallels between attention sinks and residual sinks from the perspective of outlier-driven rescaling.
Both phenomena arise at normalization layers: softmax for attention sinks and RMSNorm for residual sinks. 
They serve as modulators that control the scale of non-outlier components. Although they manifest differently—attention sinks are token-specific while residual sinks are dimension-specific—their functional roles are analogous. 
Both enable rescaling of downstream representations. 
Critically, both types of outliers are actively suppressed after fulfilling their rescaling function. 
This is evidenced by small value vector norms for attention sinks and small RMSNorm affine weights for residual sinks. 
Architectural modifications that remove normalization eliminate these outliers but degrade performance or stability. 
Conversely, explicitly providing alternative rescaling pathways, such as Gated Attention or GatedNorm, effectively reduces reliance on outliers while preserving or even enhancing model performance.


\subsection{Efficiency Analysis of GatedNorm}
\label{app:efficiency}

We evaluate the end-to-end training overhead of our method on an 8-layer dense transformer using the ZeRO-1 optimizer configuration in Megatron-LM~\citep{shoeybi2019megatron}. The hidden dimension is varied while the low-rank dimension of the gating mechanism is fixed to 16. For the PreAffine variant, we implement a custom fused kernel in Triton to ensure efficient execution.


\begin{table}[h]
\centering
\caption{End-to-end training overhead of GatedNorm as a function of hidden dimension (rank fixed at 16).}
\label{tab:overhead}
\begin{tabular}{cc}
\toprule
Hidden Size & Relative Overhead \\
\midrule
2048 & 8.1\% \\
4096 & 5.9\% \\
8192 & 3.6\% \\
\bottomrule
\end{tabular}
\end{table}

As shown, the relative overhead decreases rapidly with increasing model scale. This trend is explained by two factors. 
First, the computational cost of GatedNorm scales linearly with the hidden dimension and depends only weakly (via a constant factor) on the low-rank dimension, whereas the dominant GEMM operations in attention and FFN layers scale quadratically with the hidden size.
Consequently, the fraction of total compute attributable to GatedNorm becomes increasingly negligible at larger scales.

Second, kernel launch overhead is more significant for smaller hidden dimensions. In such cases, GatedNorm consists of multiple lightweight kernels whose launch latency can create execution bubbles and reduce hardware utilization. 
As the hidden dimension grows, the per-kernel workload increases sufficiently to amortize launch costs, improving pipeline efficiency and further reducing relative overhead.

Third, in MoE settings, the relative overhead of GatedNorm becomes even smaller. 
MoE models incur substantial communication and routing costs that dominate the training step time. 
Since GatedNorm introduces only lightweight, local computation, its contribution to the total step time is further diluted in this regime, resulting in lower relative overhead compared to dense models of similar scale.

\subsection{Experimental Setup}
\label{sec:detailed_setup}

\begin{table}[h]
\centering
\caption{Architectural specifications of the target LLMs used in our experiments. The 7B-A2B and 24B-A3B model are MoEs. All models use a head dimension of 256. Embedding weights are tied in the dense models but not in the MoE model.}
\label{tab:architectures}
\begin{tabular}{lccc}
\toprule
\textbf{Model} & \textbf{2B} & \textbf{7B-A2B} & \textbf{24B-A3B} \\
\midrule
Layers & 28 & 48 & 24 \\
Softmax Attention Interval & - & 4 & 4 \\
Softmax Attention Query Heads & 8 & 8 & 16 \\
Softmax Attention Key / Value Heads & 2 & 1 & 2 \\
Softmax Attention Head Dimension & 256 & 256 & 256 \\
Linear Attention Head Dimension & - & 128 & 128 \\
Linear Attention Value Head & - & 16 & 32 \\
Linear Attention Query / Key Head & - & 8 & 16 \\
Tie Embedding & Yes & No & No \\
Hidden Size & 2048 & 2048 & 2048 \\
FFN Size & 6144 & 384 & 512 \\
Number of Experts & -- & 128 & 256 \\
Number of Shared Experts & -- & 1 & 1 \\
Top-$k$ & -- & 6 & 8 \\
\bottomrule
\end{tabular}
\end{table}

\subsubsection{Scaling Setups}
\label{app:scaling_set}

\textbf{Model Architecture}
We evaluate our methods on a suite of large language models with diverse architectures, including both dense and MoE variants. 
The architectural specifications are summarized in Table~\ref{tab:architectures}. 
All models use a head dimension of 256 for softmax attention and share the same hidden size of 2048. The 2B model is a standard dense Transformer. 
In contrast, the 7B-A2B and 24B-A3B models are MoE architectures that combine linear and softmax attention in a hybrid configuration: softmax attention is applied every 4 layers, while linear attention is used in the remaining layers.

The MoE models employ a large number of experts (128 for 7B-A2B and 256 for 24B-A3B), with top-$k$ routing ($k=6$ and $k=8$, respectively) and one shared expert to ensure baseline capacity. 
They are trained with global load balance loss~\citep{global_balance}.
Notably, embedding weights are tied in the dense 2B model but untied in the MoE models, following common practice for large sparse architectures. The FFN expansion ratios differ significantly: the dense model uses a wide FFN (6144-dimensional), whereas the MoE models use much smaller per-expert FFNs (384 and 512, respectively), compensated by expert parallelism. 
Linear attention heads use a reduced dimensionality (128) and separate query/key and value projections, as detailed in the table.

\textbf{Evaluation}
We evaluate model performance across a broad set of benchmarks spanning knowledge, reasoning, STEM, code generation, and multilingual capabilities. 
Specifically, we report results on MMLU-Redux and MMLU-Pro for general knowledge, SuperGPQA and GPQA-Diamond for expert-level scientific reasoning, GSM8K and MATH for mathematical problem solving, CruxEval, MultiPL-E, and MBPP for code generation, and MMMLU and MGSM for multilingual understanding.

MMLU-Redux~\cite{gema2024are} is a refined version of the original MMLU benchmark with improved question quality and reduced ambiguity. 
MMLU-Pro~\cite{wang2024mmlu} extends this with more challenging, multi-hop questions requiring deeper reasoning. 
SuperGPQA~\cite{du2025supergpqa} and GPQA-Diamond~\cite{reinhardt2023gpqa} consist of expert-written, graduate-level scientific questions designed to assess advanced domain knowledge. 
GSM8K~\cite{cobbe2021training} and MATH~\cite{hendrycks2021measuring} evaluate grade-school and advanced mathematical reasoning, respectively. 
CruxEval~\cite{gu2024cruxeval} tests code generation via input-output specification completion, while MultiPL-E~\cite{cassano2023multipl} and MBPP~\cite{austen2021mbpp} assess cross-language and beginner-level Python programming. 
MMMLU~\cite{openai2024mmmlu} and MGSM~\cite{shi2022language} are multilingual extensions of MMLU and GSM8K, covering dozens of languages to evaluate cross-lingual transfer.

\subsubsection{Quantization Settings}
\label{app:quant}

\textbf{Unified Optimization Strategy.} To mitigate activation outliers, we apply \textit{SmoothQuant}~\citep{xiao2023smoothquant} as a universal pre-processing step. A calibration set of 4096 sequences is used solely to compute per-channel smoothing factors, explicitly migrating quantization difficulty from activations to weights.

\textbf{Quantization Configurations.} Building on this smoothed baseline, we evaluate two hardware-aligned formats:

\begin{itemize}[leftmargin=*, noitemsep, topsep=2pt, parsep=2pt]
    \item \textbf{FP8 (W8A8):} We employ the \texttt{E4M3} format. Weights are quantized using a $128 \times 128$ \textit{per-block} scaling strategy, while activations utilize dynamic \textit{per-token} quantization.
    
    \item \textbf{FP4 (W4A4):} We utilize the NVIDIA FP4~\citep{nvfp4} format with \textit{hierarchical two-stage scaling}. Weights are grouped into blocks of 16, where a shared floating-point scale (1st stage) normalizes the range before 4-bit mapping (2nd stage). 
    For activations, we modify the 1st-stage scaling from static to \textit{dynamic per-token} to preserve fidelity.
\end{itemize}

\subsection{More Visualization Results}

\subsubsection{RMSNorm Visualization}
\label{app:rmsnorm_weight}

\begin{figure}[ht]
    \centering
    \includegraphics[width=0.9\linewidth]{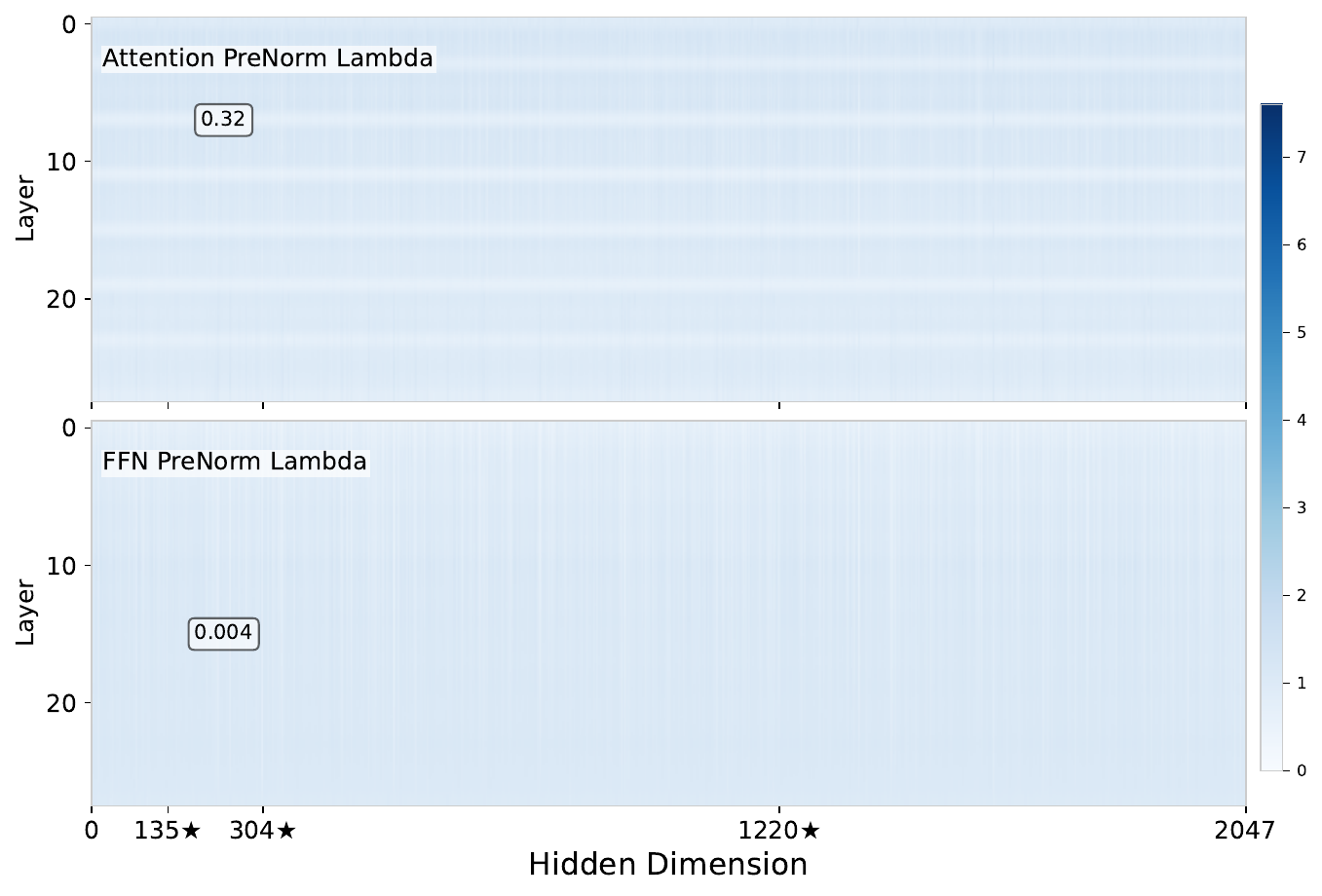}
    \caption{RMSNorm weights for baseline.}
    \label{fig:rmsnorm_baseline}
\end{figure}

\begin{figure}[ht]
    \centering
    \includegraphics[width=0.9\linewidth]{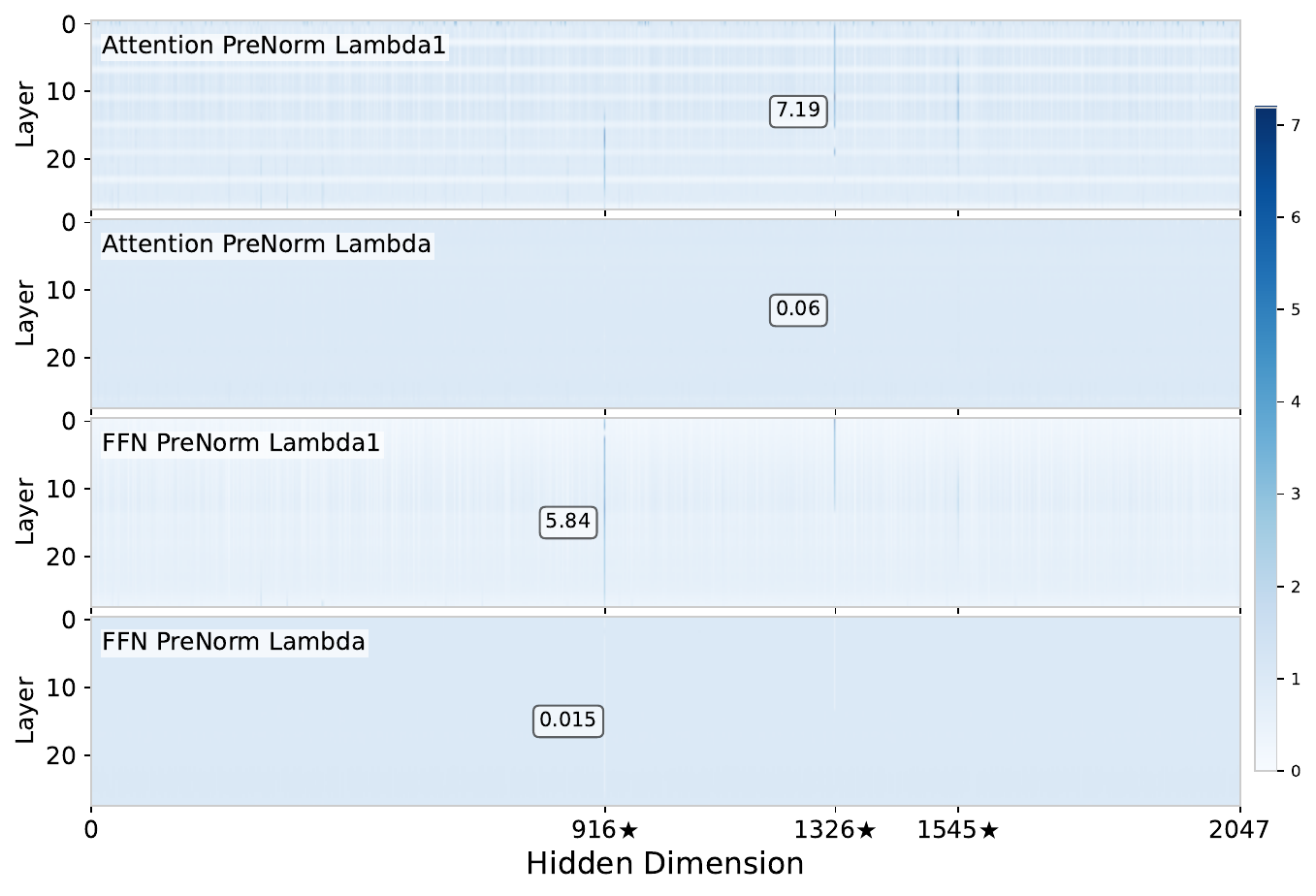}
    \caption{RMSNorm weights for PreAffineRMSNorm.}
    \label{fig:rmsnorm_preaffine}
\end{figure}

\begin{figure}[ht]
    \centering
    \includegraphics[width=0.9\linewidth]{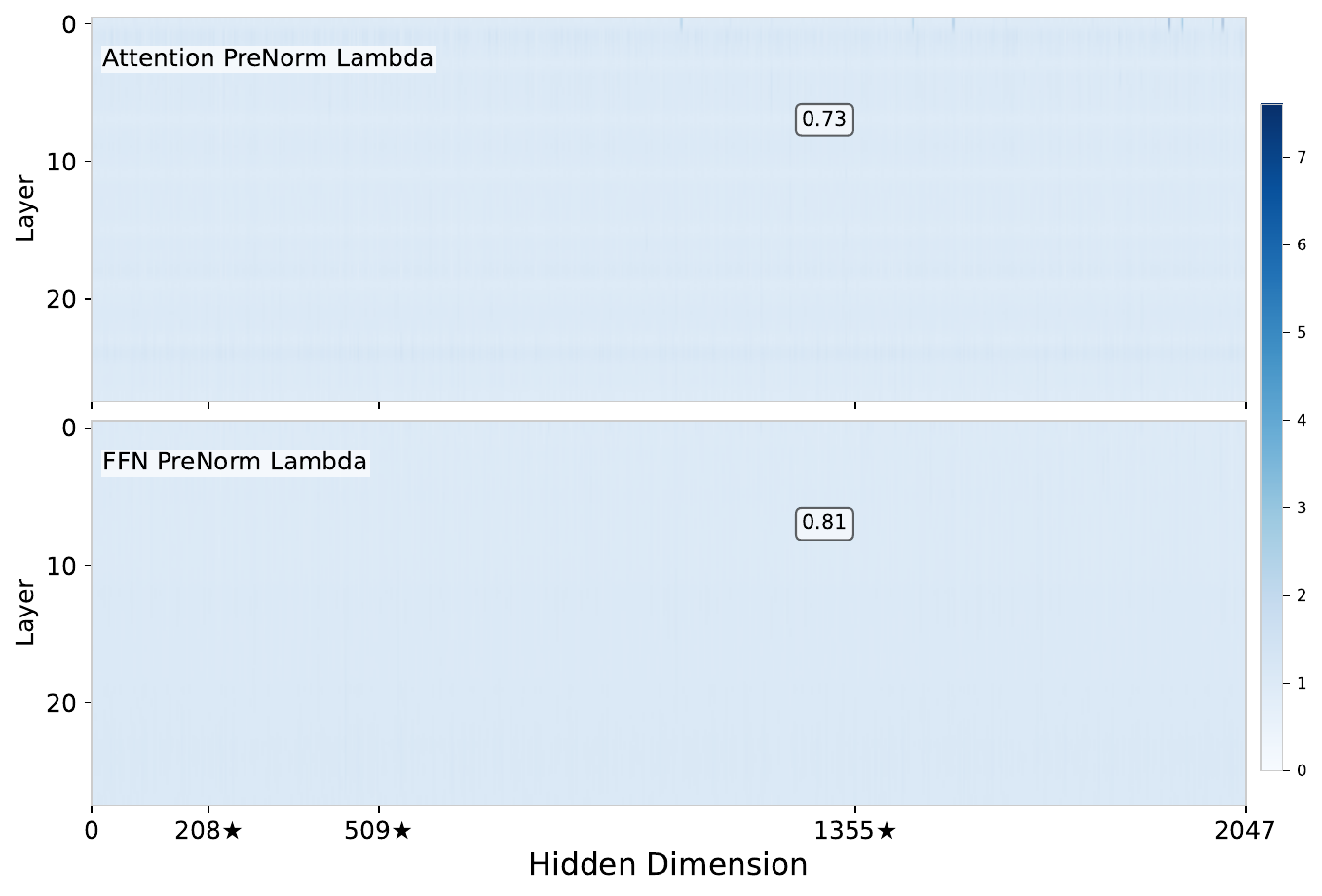}
    \caption{RMSNorm weights for GatedRMSNorm. 
    Weights in most dimensions are near 1, while the largest deviation to 1 is 0.73. }
    \label{fig:rmsnorm_gated}
\end{figure}

\subsubsection{Outliers for Other Models}

\begin{figure}[ht]
    \centering
    \includegraphics[width=0.9\linewidth]{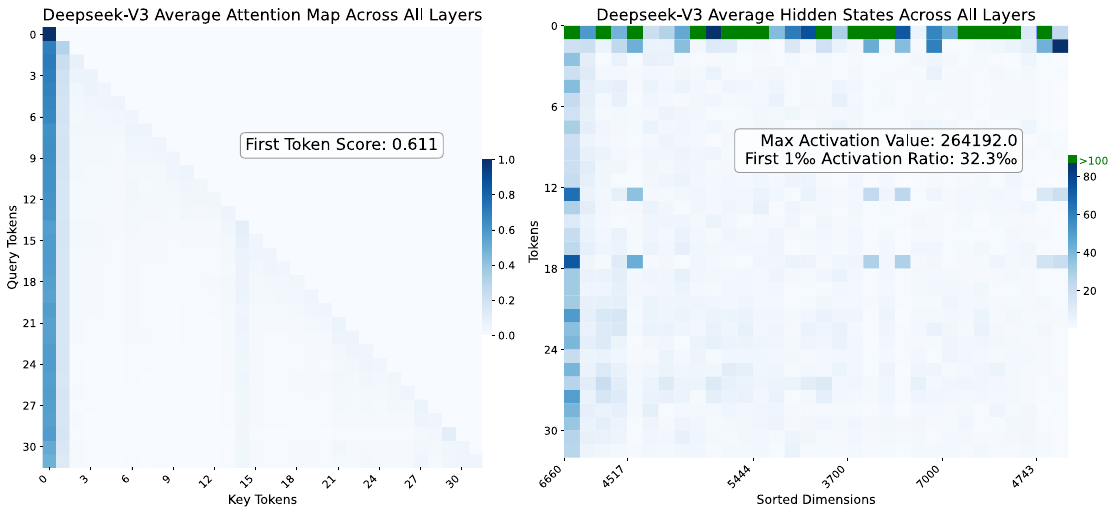}
    \caption{Outliers for Deepseek-V3. The activation pattern of <begin of sentence> token is different from other tokens. Attention sink and residual sink both exists.}
    \label{fig:reduce_sink_dpsk}
\end{figure}

\end{document}